\documentclass[10pt,twocolumn,letterpaper]{article}

\usepackage{iccv}
\usepackage{times}
\usepackage{epsfig}
\usepackage{graphicx}
\usepackage{amsmath}
\usepackage{amssymb}

\usepackage[pagebackref=true,breaklinks=true,letterpaper=true,colorlinks,bookmarks=false]{hyperref}

\iccvfinalcopy 

\def\iccvPaperID{753} 

\ificcvfinal\fi
\begin{document}

\title{Learning to Synthesize a 4D RGBD Light Field from a Single Image}

\author{Pratul P. Srinivasan$^1$, Tongzhou Wang$^1$, Ashwin Sreelal$^1$, Ravi Ramamoorthi$^2$, Ren Ng$^1$\\
$^1$University of California, Berkeley \qquad $^2$University of California, San Diego\\
}

\maketitle
\thispagestyle{empty}

\begin{abstract}

We present a machine learning algorithm that takes as input a 2D RGB image and synthesizes a 4D RGBD light field (color and depth of the scene in each ray direction). For training, we introduce the largest public light field dataset, consisting of over 3300 plenoptic camera light fields of scenes containing flowers and plants. Our synthesis pipeline consists of a convolutional neural network (CNN) that estimates scene geometry, a stage that renders a Lambertian light field using that geometry, and a second CNN that predicts occluded rays and non-Lambertian effects. Our algorithm builds on recent view synthesis methods, but is unique in predicting RGBD for each light field ray and improving unsupervised single image depth estimation by enforcing consistency of ray depths that should intersect the same scene point.

\end{abstract}
\vspace{-0.2in}

\begin{figure*}
\begin{center}
\newcommand{\width}{1.0\linewidth}
\includegraphics[width=\width]{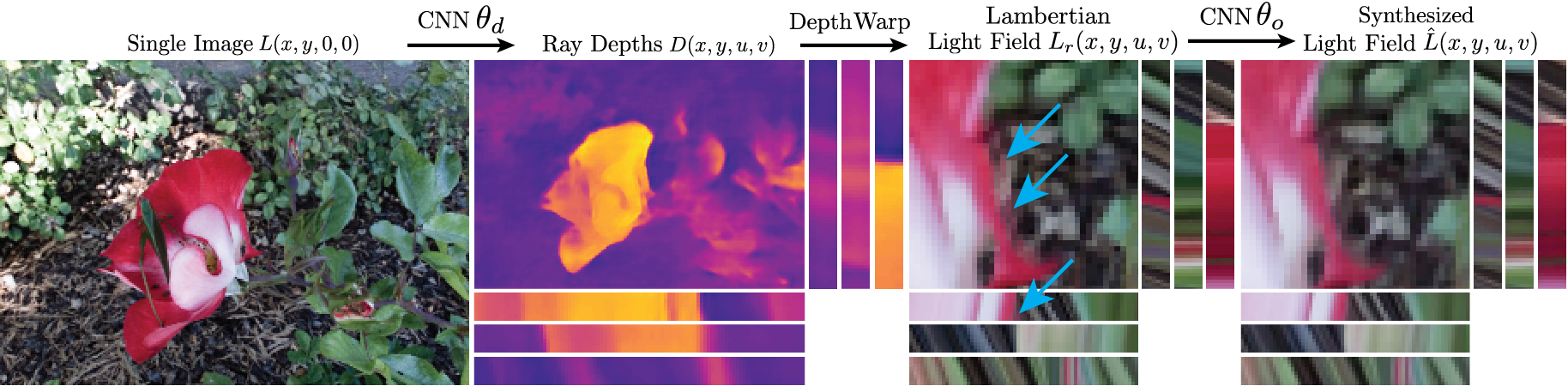} 
\caption{We propose a CNN framework that factors the light field synthesis problem into estimating depths for each ray in the light field, rendering a Lambertian approximation to the light field, and refining this approximation by predicting occluded rays and non-Lambertian effects (incorrect rays that are refined, in this case red rays that should be the color of the background instead of the flower, are marked with blue arrows). We train this network end-to-end by minimizing the reconstruction errors of the Lambertian and predicted light fields, along with a novel physically-based depth regularization. We demonstrate that we can predict convincing 4D light fields and ray depths from a single 2D image. We visualize synthesized light fields as a predicted corner view along with epipolar slices in both the $u$ and $v$ directions of different spatial segments. Please view our supplementary video for compelling animations of our light fields and ray depths.}
\label{fig:overview}
\end{center}
\vspace{-0.2in}
\end{figure*}

\section{Introduction}

We focus on a problem that we call ``local light field synthesis", which we define as the promotion of a single photograph to a plenoptic camera light field. One can think of this as expansion from a single view to a dense 2D patch of views. We argue that local light field synthesis is a core visual computing problem with high potential impact. First, it would bring light field benefits such as synthetic apertures and refocusing to everyday photography. Furthermore, local light field synthesis would systematically lower the sampling rate of photographs needed to capture large baseline light fields, by ``filling the gap" between discrete viewpoints. This is a path towards making light field capture for virtual and augmented reality (VR and AR) practical. In this work, we hope to convince the community that local light field synthesis is actually a tractable problem. 

From an alternative perspective, the light field synthesis task can be used as an unsupervised learning framework for estimating scene geometry from a single image. Without any ground-truth geometry for training, we can learn to estimate the geometry that minimizes the difference between the light field rendered with that geometry and the ground-truth light field.

Light field synthesis is a severely ill-posed problem, since the goal is to reconstruct a 4D light field given just a single image, which can be interpreted as a 2D slice of the 4D light field. To alleviate this, we use a machine learning approach that is able to utilize prior knowledge of natural light fields. In this paper, we focus on scenes of flowers and plants, because they contain interesting and complex occlusions as well as a wide range of relative depths. Our specific contributions are the introduction of the largest available light field dataset, the prediction of 4D ray depths with a novel depth consistency regularization to improve unsupervised depth estimation, and a learning framework to synthesize a light field from a single image.

\vspace{-0.1in}
\paragraph{Light Field Dataset}
We collect the largest available light field dataset (Sec.~\ref{sec:dataset}), contaning 3343 light fields of flowers and plants, taken with the Lytro Illum camera. Our dataset limits us to synthesizing light fields with camera-scale baselines, but we note that our model can generalize to light fields of any scene and baseline given the appropriate datasets.

\vspace{-0.1in}
\paragraph{Ray Depths and Regularization}
Current view synthesis methods generate each view separately. Instead, we propose to concurrently predict the entire 4D light field by estimating a separate depth map for each viewpoint, which is equivalent to estimating a depth for each ray in the 4D light field (Sec.~\ref{sec:depth}). We introduce a novel physically-based regularization that encourages the predicted depth maps to be consistent across viewpoints, alleviating typical problems that arise in depths created by view synthesis (Fig.~\ref{fig:reg_results}). We demonstrate that our algorithm can predict depths from a single image that are comparable or better than depths estimated by a state-of-the-art physically-based non-learning method that uses the entire light field~\cite{Jeon15} (Fig.~\ref{fig:depth_results}).

\vspace{-0.1in}
\paragraph{CNN Framework}
We create and study an end-to-end convolutional neural network (CNN) framework, visualized in Fig.~\ref{fig:overview}, that factorizes the light field synthesis problem into the subproblems of estimating scene depths for every ray (Fig.~\ref{fig:depth_results}, Sec.~\ref{sec:depth}) (we use depth and disparity interchangeably, since they are closely related in structured light fields), rendering a Lambertian light field (Sec.~\ref{sec:render}), and predicting occluded rays and non-Lambertian effects (Sec.~\ref{sec:occlusions}). This makes the learning process more tractable and allows us to estimate scene depths, even though our network is trained without any access to the ground truth depths. Finally, we demonstrate that it is possible to synthesize high-quality ray depths and light fields of flowers and plants from a single image (Fig.~\ref{fig:overview}, Fig.~\ref{fig:depth_results}, Fig.~\ref{fig:results}, Fig.~\ref{fig:cell_phone}, Sec.~\ref{sec:results}).

\section{Related Work}

\paragraph{Light Fields}
The 4D light field~\cite{Lippmann1908} is the total spatio-angular distribution of light rays passing through a region of free space. Previous work has demonstrated exciting applications of light fields, including rendering images from new viewpoints~\cite{Levoy96}, changing the focus and depth-of-field of photographs after capture~\cite{Ng05}, correcting lens aberrations~\cite{Ng06}, and estimating scene flow~\cite{Srinivasan15}.

\paragraph{View Synthesis from Light Fields}
Early work on light field rendering~\cite{Levoy96} captures a densely-sampled 4D light field of a scene, and renders images from new viewpoints as 2D slices of the light field. Closely related work on the Lumigraph~\cite{Gortler96} uses approximate geometry information to refine the rendered slices. The unstructured Lumigraph rendering framework~\cite{Buehler01} extends these approaches to use a set of unstructured (not axis-aligned in the angular dimensions) 2D slices of the light field. In contrast to these pioneering works which capture many 2D slices of the light field to render new views, we propose to synthesize a dense sampling of new views from just a single slice of the light field.

\paragraph{View Synthesis without Geometry Estimation}
Alternative approaches synthesize images from new viewpoints without explicitly estimating geometry. The work of Shi \etal~\cite{Shi15} uses the observation that light fields are sparse in the continuous Fourier domain to reconstruct a full light field from a carefully-constructed 2D collection of views. Didyk \etal~\cite{Didyk13} and Zhang \etal~\cite{Zhang15} reconstruct 4D light fields from pairs of 2D slices using phase-based approaches.

Recent works have trained CNNs to synthesize slices of the light field that have dramatically different viewpoints than the input slices. Tatarchenko \etal~\cite{Tatarchenko16} and Yang \etal~\cite{Yang15} train CNNs to regress from a single input 2D view to another 2D view, given the desired camera rotation. The exciting work of Zhou \etal~\cite{Zhou16} predicts a flow field that rearranges pixels from the input views to synthesize novel views that are sharper than directly regressing to pixel values. These methods are trained on synthetic images rendered from large databases of 3D models of objects such as cars and chairs~\cite{shapenet15}, while we train on real light fields. Additionally, they are not able to explicitly take advantage of geometry because they attempt to synthesize views at arbitrary rotations with potentially no shared geometry between the input and target views. We instead focus on the problem of synthesizing a dense sampling of views around the input view, so we can explicitly estimate geometry to produce higher quality results.

\paragraph{View Synthesis by Geometry Estimation}
Other methods perform view interpolation by first estimating geometry from input 2D slices of the light field, and then warping the input views to reconstruct new views. These include view interpolation algorithms~\cite{Chaurasia13, Goesele10} which use wider baseline unstructured stereo pairs to estimate geometry using multi-view stereo algorithms. 

More recently, CNN-based view synthesis methods been proposed, starting with the inspiring DeepStereo method that uses unstructured images from Google's Street View~\cite{Flynn16} to synthesize new views. This idea has been extended to view interpolation for light fields given 4 corner views~\cite{Kalantari16}, and the prediction of one image from a stereo pair given the other image~\cite{Garg16,Godard17,Xie16}.

We take inspiration from the geometry-based view synthesis algorithms discussed above, and also predict geometry to warp an input view to novel views. However, unlike previous methods, we synthesize an entire 4D light field from just a single image. Furthermore, we synthesize all views and corresponding depths at once, as opposed to the typical strategy of predicting a single 2D view at a time, and leverage this to produce better depth estimations. 

\paragraph{3D Representation Inference from a Single Image}
Instead of synthesizing new imagery, many excellent works address the general inverse rendering problem of inferring the scene properties that produce an observed 2D image. The influential algorithm of Barron and Malik~\cite{Barron15} solves an optimization problem with priors on reflectance, shape, and illumination to infer these from a single image. Other interesting works~\cite{Eigen15,Saxena08} focus on inferring just the 3D structure of the scene, and train on ground-truth geometry captured with 3D scanners or the Microsoft Kinect. A number of exciting works extend this idea to infer a 3D voxel~\cite{Choy16,Girdhar16,Wu16} or point set~\cite{Fan17} representation from a synthetic 2D image by training CNNs on large databases of 3D CAD models. Finally, recent methods~\cite{Rezende16,Tulsiani17,Yan16} learn to infer 3D voxel grids from a 2D image without any 3D supervision by using a rendering or projection layer within the network and minimizing the error of the rendered view. Our work is closely related to unsupervised 3D representation learning methods, but we represent geometry as 4D ray depths instead of voxels, and train on real light fields instead of views from synthetic 3D models of single objects. 

\section{Light Field Synthesis}

\begin{figure}
\begin{center}
\newcommand{\width}{1.0\linewidth}
\includegraphics[width=\width]{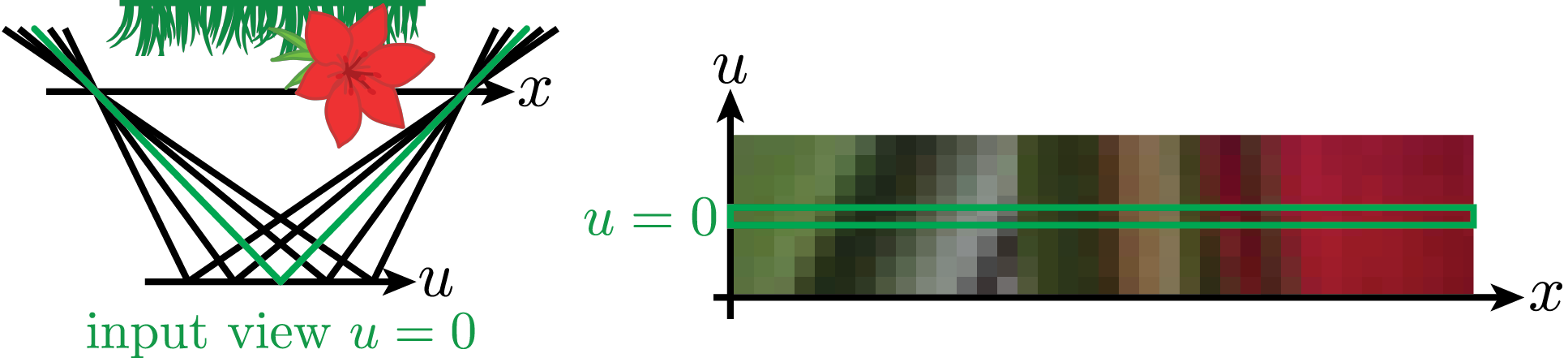} 
\caption{Two equivalent interpretations of the local light field synthesis problem. Left: Given an input image of a scene, with the field-of-view marked in green, our goal is to synthesize a dense grid of surrounding views, with field-of-views marked in black. The $u$ dimension represents the center-of-projection of each virtual viewpoint, and the $x$ axis represents the optical conjugate of the sensor plane. Right: Given an input image, which is a 1D slice of the 2D flatland light field (2D slice of the full 4D light field), our goal is to synthesize the entire light field. In our light field parameterization, vertical lines correspond to points in focus, and lines at a slope of 45 degrees correspond to points at the farthest distance that is within the depth of field of each sub-aperture image.}
\label{fig:formulation}
\end{center}
\vspace{-0.2in}
\end{figure}

Given an image from a single viewpoint, our goal is to synthesize views from a densely-sampled grid around the input view. This is equivalent to synthesizing a 4D light field, given a central 2D slice of the light field, and both of these interpretations are visualized in Fig.~\ref{fig:formulation}. We do this by learning to approximate a function $f$:
\begin{equation}
\label{eq:f}
\hat{L}(\mathbf{x},\mathbf{u})=f(L(\mathbf{x},\mathbf{0}))
\end{equation} 
where $\hat{L}$ is the predicted light field, $\mathbf{x}$ is spatial coordinate $(x,y)$, $\mathbf{u}$ is angular coordinate $(u,v)$, and $L(\mathbf{x},\mathbf{u})$ is the ground-truth light field, with input central view $L(\mathbf{x},\mathbf{0})$.

Light field synthesis is severely ill-posed, but certain redundancies in the light field as well as prior knowledge of scene statistics enable us to infer other slices of the light field from just a single 2D slice. Figure~\ref{fig:formulation} illustrates that scene points at a specific depth lie along lines with corresponding slopes in the light field. Furthermore, the colors along these lines are constant for Lambertian reflectance, and only change due to occlusions or non-Lambertian reflectance effects. 

We factorize the problem of light field synthesis into the subproblems of estimating the depth at each coordinate $(\mathbf{x},\mathbf{u})$ in the light field, rendering a Lambertian approximation of the light field using the input image and these estimated depths, and finally predicting occluded rays and non-Lambertian effects. This amounts to factorizing the function $f$ in Eq.~\ref{eq:f} into a composition of 3 functions: $d$ to estimate ray depths, $r$ to render the approximate light field from the depths and central 2D slice, and $o$ to predict occluded rays and non-Lambertian effects from the approximate light field and predicted depths:
\begin{equation}
\label{eq:factor}
\begin{split}
D(\mathbf{x},\mathbf{u})=&d(L(\mathbf{x},\mathbf{0}))\\
L_r(\mathbf{x},\mathbf{u})=&r(L(\mathbf{x},\mathbf{0}),D(\mathbf{x},\mathbf{u}))\\
\hat{L}(\mathbf{x},\mathbf{u})=&o(L_r(\mathbf{x},\mathbf{u}),D(\mathbf{x},\mathbf{u}))
\end{split}
\end{equation}
where $D(\mathbf{x},\mathbf{u})$ represents predicted ray depths, and $L_r$ represents the rendered Lambertian approximate light field. This factorization lets the network learn to estimate scene depths from a single image in an unsupervised manner. 

The rendering function $r$ (Sec.~\ref{sec:render}) is physically-based, while the depth estimation function $d$ (Sec.~\ref{sec:depth}) and occlusion prediction function $o$ (Sec.~\ref{sec:occlusions}) are both structured as CNNs, due to their state-of-the-art performance across many function approximation problems in computer vision. The CNN parameters are learned end-to-end by minimizing the sum of the reconstruction error of the Lambertian approximate light field, the reconstruction error of the predicted light field, and regularization losses for the predicted depths, for all training tuples:
\begin{equation}
\label{eq:loss}
\begin{split}
\min_{\theta_d,\theta_o}\sum_{\mathcal{S}}&\big[||L_r-L||_1+||\hat{L}-L||_1\\
+&\lambda_c\psi_c(D)+\lambda_{tv}\psi_{tv}(D)\big]
\end{split}
\end{equation}
where $\theta_d$ and $\theta_o$ are the parameters for the depth estimation and occlusion prediction networks. $\psi_c$ and $\psi_{tv}$ are consistency and total variation regularization losses for the predicted ray depths, discussed below in Sec.~\ref{sec:depth}. $\mathcal{S}$ is the set of all training tuples, each consisting of an input central view $L(\mathbf{x},\mathbf{0})$ and ground truth light field $L(\mathbf{x},\mathbf{u})$. 

We include the reconstruction errors for both the Lambertian light field and the predicted light field in our loss to prevent the occlusion prediction network from attempting to learn the full light field prediction function by itself, which would prevent the depth estimation network from properly learning a depth estimation function. 

\begin{figure*}
\begin{center}
\newcommand{\width}{1.0\linewidth}
\includegraphics[width=\width]{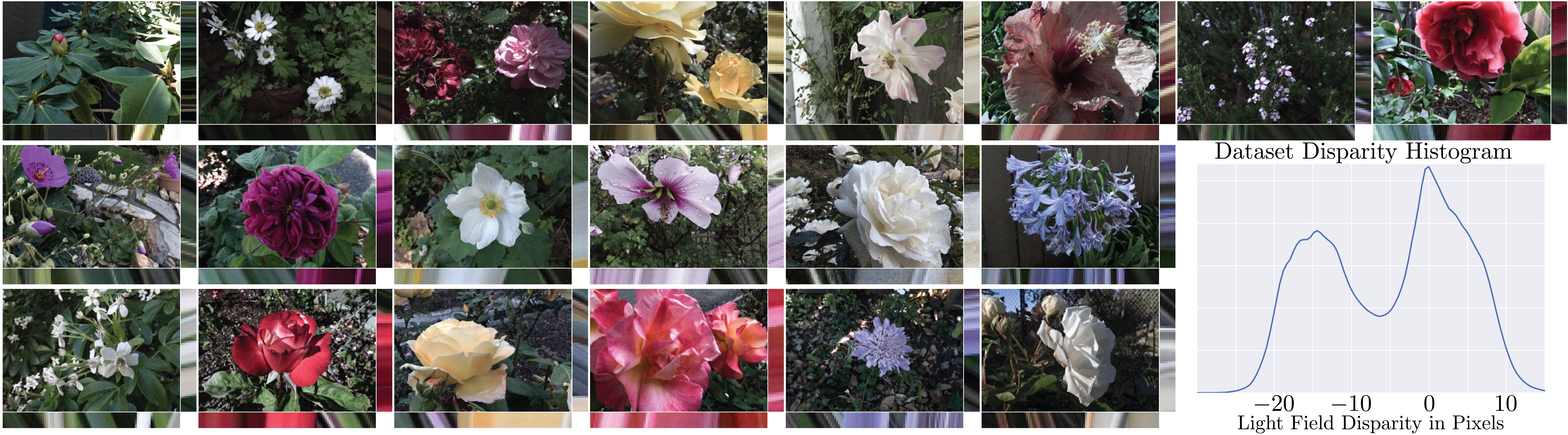} 
\caption{We introduce the largest available light field dataset, containing 3343 light fields of scenes of flowers and plants captured with the Lytro Illum camera in various locations and lighting settings. These light fields contain complex occlusions and wide ranges of relative depths, as visualized in the example epipolar slices. No ground truth depths are available, so we use our algorithm to predict a histogram of disparities in the dataset to demonstrate the rich depth complexity in our dataset. We will make this dataset available upon publication.}
\label{fig:dataset}
\end{center}
\vspace{-0.2in}
\end{figure*}

\begin{figure}
\begin{center}
\newcommand{\width}{1.0\linewidth}
\includegraphics[width=\width]{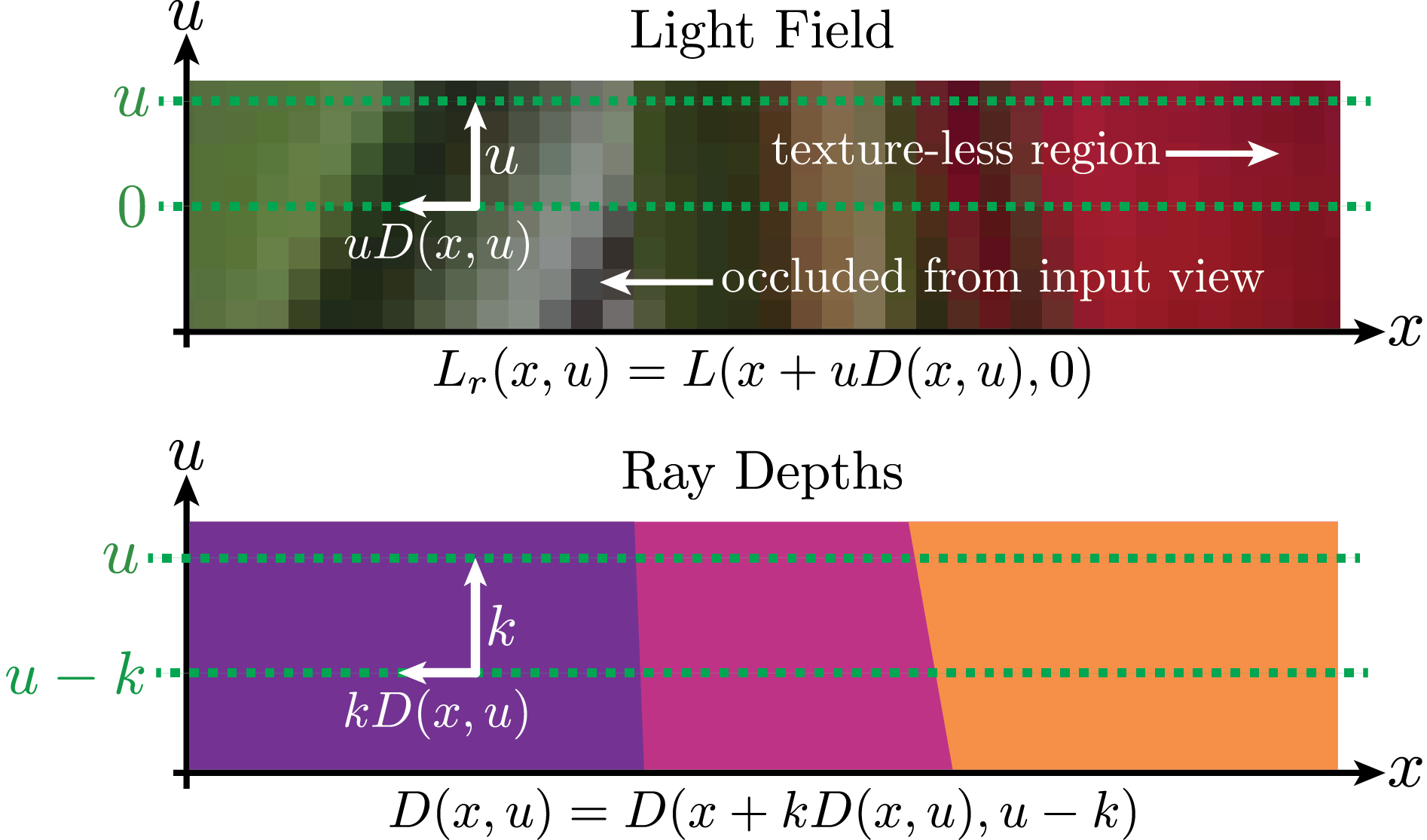} 
\caption{Top: In a Lambertian approximation of the light field, the color of a scene point is constant along the line corresponding to its depth. Given estimated disparities $D(x,u)$ and a central view $L(x,0)$, we can render the flatland light field as $L(x,u)=L(x+uD(x,u),0)$ ($D(x,u)$ is negative in this example). In white, we illustrate two prominent problems that arise when estimating depth by minimizing the reconstruction error of novel views. It is difficult to estimate the correct depth for points occluded from the input view, because warping the input view using the correct depth does not properly reconstruct the novel views. Additionally, it is difficult to estimate the correct depth in texture-less regions, because many possible depths result in the same synthesized novel views. Bottom: Analogous to the Lambertian color consistency, rays from the same scene point should have the same depth. This can be represented as $D(x,u)=D(x+kD(x,u),u-k)$ for any continuous value of $k$. We visualize ray depths using a colormap where darker colors correspond to further objects.}
\label{fig:rendering}
\end{center}
\vspace{-0.2in}
\end{figure}

\begin{figure}
\begin{center}
\newcommand{\width}{1.0\linewidth}
\includegraphics[width=\width]{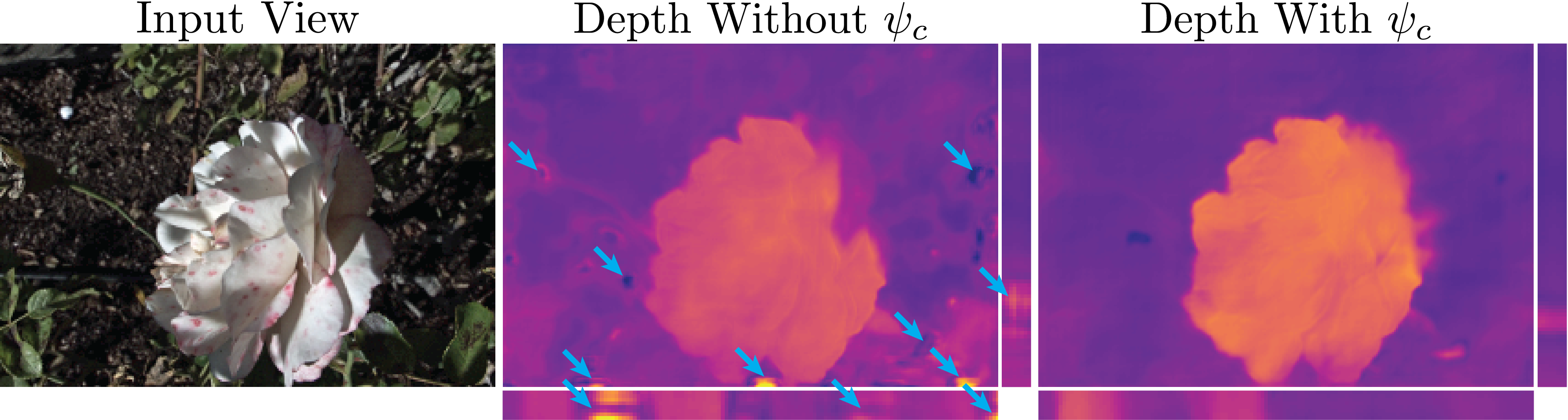} 
\caption{Our proposed phyiscally-based depth consistency regularization produces higher-quality estimated depths. Here, we visualize example sub-aperture depth maps where our novel regularization improves the estimated depths for texture-less regions. Blue arrows indicate incorrect depths and depths that are inconsistent across views, as shown in the epipolar slices.}
\label{fig:reg_results}
\end{center}
\vspace{-0.2in}
\end{figure}

\section{Light Field Dataset}
\label{sec:dataset}

To train our model, we collected 3343 light fields of flowers and plants with the Lytro Illum camera, randomly split into 3243 for training and 100 for testing. We captured all light fields using a focal length of 30 mm and $f/2$ aperture. Other camera parameters including the shutter speed, ISO, and white balance were set automatically by the camera. We decoded the sensor data from the Illum camera using the Lytro Power Tools Beta decoder, which demosaics the color sensor pattern and calibrates the lenslet locations. Each light field has 376x541 spatial samples, and 14x14 angular samples. Many of the corner angular samples lie outside the camera's aperture, so we used an 8x8 grid of angular samples in our experiments, corresponding to the angular samples that lie fully within the aperture.

This dataset includes light fields of several varieties of roses, poppies, thistles, orchids, lillies, irises, and other plants, all of which contain complex occlusions. Furthermore, these light fields were captured in various locations and times of day with different natural lighting conditions. Figure~\ref{fig:dataset} illustrates the diversity of our dataset, and the geometric complexity in our dataset can be visualized in the epipolar slices. To quantify the geometric diversity of our dataset, we compute a histogram of the disparities across the full aperture using our trained depth estimation network, since we do not have ground truth depths. The left peak of this histogram corresponds to background points, which have large negative disparities, and the right peak of the histogram corresponds to the photograph subjects (typically flowers) which are in focus and have small disparities.

We hope this dataset will be useful for future investigations into various problems including light field synthesis, single view synthesis, and unsupervised geometry learning.

\section{Synthesizing 4D Ray Depths}
\label{sec:depth}

We learn the function $d$ to predict depths by minimizing the reconstruction error of the rendered Lambertian light field, along with our novel depth regularization.

Two prominent errors arise when learning to predict depth maps by minimizing the reconstruction error of synthesized views, and we visualize these in Fig.~\ref{fig:rendering}. In texture-less regions, the depth can be incorrect and depth-based warping will still synthesize the correct image. Therefore, the minimization in Eq.~\ref{eq:loss} has no incentive to predict the correct depth. Second, depths for scene points that are occluded from the input view are also typically incorrect, because predicting the correct depth would cause the synthesized view to sample pixels from the occluder.

Incorrect depths are fine if we only care about the synthesized views. However, the quality of these depths must be improved to consider light field synthesis as an unsupervised learning algorithm to infer depth from a single 2D image. It is difficult to capture large datasets of ground-truth depths for real scenes, especially outdoors, while it is much easier to use capture scenes with a plenoptic camera. We believe that light field synthesis is a promising way to train algorithms to estimate depths from a single image, and we present a strategy to address these depth errors.

We predict depths for every ray in the light field, which is equivalent to predicting a depth map for each view. This enables us to introduce a novel regularization that encourages the predicted depths to be consistent across views and accounts for occlusions, which is a light field generalization of the left-right consistency used in methods such as~\cite{Godard17, Zitnick04}. Essentially, depths should be consistent for rays coming from the same scene points, which means that the ray depths should be consistent along lines with the same slope:
\begin{equation}
\label{eq:depth_consistency}
\begin{split}
D(\mathbf{x},\mathbf{u})=D(\mathbf{x}+\mathbf{k}D(\mathbf{x},\mathbf{u}),\mathbf{u}-\mathbf{k})
\end{split}
\end{equation}
for any continuous value of $\mathbf{k}$, as illustrated in Fig.~\ref{fig:rendering}.

To regularize the predicted depth maps, we minimize the $L_1$ norm of finite-difference gradients along these sheared lines by setting $k=1$, which both encourages the predicted depths to be consistent across views and encourages occluders to be sparse:
\begin{equation}
\label{eq:depth_reg}
\begin{split}
\psi_c(D(\mathbf{x},\mathbf{u}))=||D(\mathbf{x},\mathbf{u})-D(\mathbf{x}+D(\mathbf{x},\mathbf{u}),\mathbf{u}-1)||_1
\end{split}
\end{equation}
where $\psi_c$ is the consistency regularization loss for predicted ray depths $D(\mathbf{x},\mathbf{u})$.

Benefits of this regularization are demonstrated in Fig.~\ref{fig:reg_results}. It encourages consistent depths in texture-less areas as well as for rays occluded from the input view, because predicting the incorrect depths would result in higher gradients along sheared lines as well as new edges in the ray depths.

We additionally use total variation regularization in the spatial dimensions for the predicted depth maps, to encourage them to be sparse in the spatial gradient domain:
\begin{equation}
\label{eq:depth_reg_tv}
\begin{split}
\psi_{tv}(D(\mathbf{x},\mathbf{u}))=||\nabla_{\mathbf{x}}D(\mathbf{x},\mathbf{u})||_1
\end{split}
\end{equation}
\vspace{-0.2in}
\paragraph{Depth Estimation Network}

We model the function $d$ to estimate 4D ray depths from the input view as a CNN. We use dilated convolutions~\cite{Yu16}, which allow the receptive field of the network to increase exponentially as a function of the network depth. Hence, each of the predicted ray depths has access to the entire input image without the resolution loss caused by spatial downsampling or pooling. Every convolution layer except for the final layer consists of a 3x3 filter, followed by batch normalization~\cite{Ioffe15} and an exponential linear unit activation function (ELU)~\cite{Clevert16}. The last layer is followed by a scaled tanh activation function instead of an ELU to constrain the possible disparities to $[-16,16]$ pixels. Please refer to our supplementary material for a more detailed network architecture description.

\section{Synthesizing the 4D Light Field}
\subsection{Lambertian Light Field Rendering}
\label{sec:render}

We render an approximate Lambertian light field by using the predicted depths to warp the input view as:
\begin{equation}
\label{eq:render}
L_r(\mathbf{x},\mathbf{u})=L(\mathbf{x}+\mathbf{u}D(\mathbf{x},\mathbf{u}),\mathbf{0})
\end{equation}
where $D(\mathbf{x},\mathbf{u})$ is the predicted depth for each ray in the light field. Figure~\ref{fig:rendering} illustrates this relationship.

This formulation amounts to using the predicted depths for each ray to render the 4D light field by sampling the input central view image. Since our depth regularization encourages the ray depths to be consistent across views, this effectively encourages different views of the same scene point to sample the same pixel in the input view, resulting in a Lambertian approximation to the light field.

\subsection{Occlusions and Non-Lambertian Effects}
\label{sec:occlusions}

Although predicting a depth for each ray, combined with our depth regularization, allows the network to learn to model occlusions, the Lambertian light fields rendered using these depths are not able to correctly synthesize the values of rays that are occluded from the input view, as demonstrated in Fig.~\ref{fig:overview}. Furthermore, this depth-based rendering is not able to accurately predict non-Lambertian effects.

We model the function $o$ to predict occluded rays and non-Lambertian effects as a residual block~\cite{He16}:
\begin{equation}
\label{eq:occlusions}
o(L_r(\mathbf{x},\mathbf{u}),D(\mathbf{x},\mathbf{u}))=\tilde{o}(L_r(\mathbf{x},\mathbf{u}),D(\mathbf{x},\mathbf{u}))+L_r(\mathbf{x},\mathbf{u})
\end{equation}
where $\tilde{o}$ is modeled as a 3D CNN. We stack all sub-aperture images along one dimension and use a 3D CNN so each filter has access to every 2D view. This 3D CNN predicts a residual that, when added to the approximate Lambertian light field, best predicts the training example true light fields. Structuring this network as a residual block ensures that decreases in the loss are driven by correctly predicting occluded rays and non-Lambertian effects. Additionally, by providing the predicted depths, this network has the information necessary to understand which rays in the approximate light field are incorrect due to occlusions. Figure~\ref{fig:quantitative} quantitatively demonstrates that this network improves the reconstruction error of the synthesized light fields.

We simply concatenate the estimated depths to the Lambertian approximate light field as the input to a 3D CNN that contains 5 layers of 3D convolutions with 3x3x3 filters (height x width x color channels), batch normalization, and ELU activation functions. The last convolutional layer is followed by a tanh activation function instead of an ELU, to constrain the values in the predicted light field to $[-1,1]$. Please refer to our supplementary material for a more detailed network architecture description.

\subsection{Training}
\label{sec:training}

We generate training examples by randomly selecting 192x192x8x8 crops from the training light fields, and spatially downsampling them to 96x96x8x8. We use bilinear interpolation to sample the input view for the Lambertian depth-based rendering, so our network is fully differentiable. We train our network end-to-end using the first-order Adam optimization algorithm~\cite{Kingma15} with default parameters $\beta_1=0.9$, $\beta_2=0.999$, $\epsilon=1e-08$, a learning rate of $0.001$, a minibatch size of 4 examples, and depth regularization parameters $\lambda_c=0.005$ and $\lambda_{tv}=0.01$. 

\section{Results}
\label{sec:results}

We validate our light field synthesis algorithm using our testing dataset, and demonstrate that we are able to synthesize compelling 4D ray depths and light fields with complex occlusions and relative depths. It is difficult to fully appreciate 4D light fields in a paper format, so we request readers to view our supplementary video for animations that fully convey the quality of our synthesized light fields. No other methods have attempted to synthesize a full 4D light field or 4D ray depths from a single 2D image, so we separately compare our estimated depths to a state-of-the-art light field depth estimation algorithm and our synthesized light fields to a state-of-the-art view synthesis method.

\begin{figure}
\begin{center}
\newcommand{\width}{1.0\linewidth}
\includegraphics[width=\width]{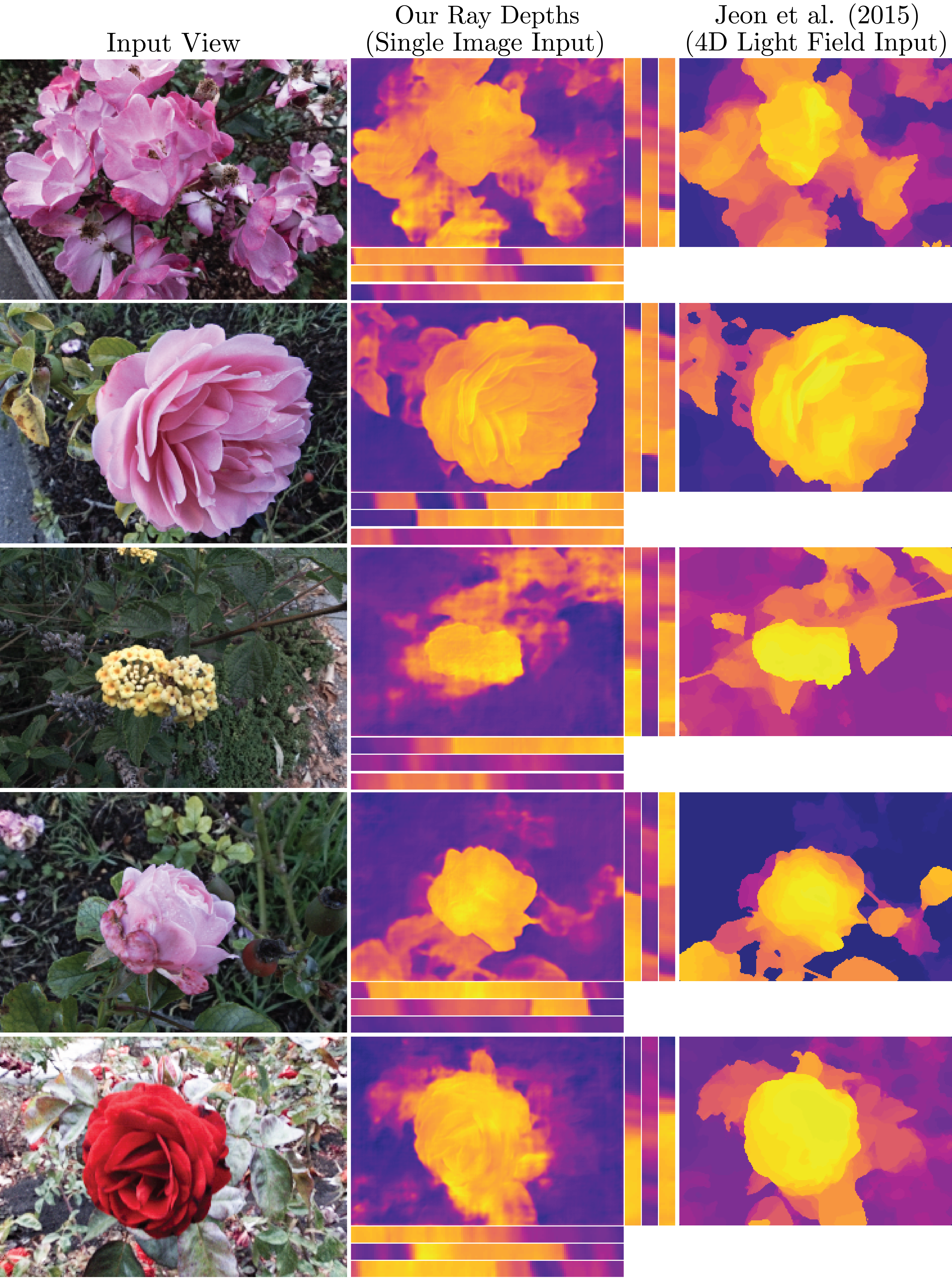} 
\caption{We validate our ray depths against the state-of-the-art light field depth estimation. We give Jeon \etal~\cite{Jeon15} a distinct advantage by providing them a ground-truth 4D light field to predict 2D depths, while we use a single 2D image to predict 4D depths. Our estimated depths are comprable, and in some cases superior, to their estimated depths, as shown by the detailed varying depths of the flower petals, leaves, and fine stem structures.}
\label{fig:depth_results}
\end{center}
\vspace{-0.2in}
\end{figure}

\vspace{-0.1in}
\paragraph{Depth Evaluation}
We compare our predicted depths to Jeon \etal~\cite{Jeon15}, which is a physically-based non-learning depth estimation technique. Note that their algorithm uses the entire ground-truth light field to estimate a 2D depth map, while our algorithm estimates 4D ray depths from a single 2D image. Figure~\ref{fig:depth_results} qualitatively demonstrates that our unsupervised depth estimation algorithm produces results that are comprable to Jeon \etal, and even more detailed in many cases. 

\vspace{-0.1in}
\paragraph{Synthesized Light Field Evaluation}
We compare our synthesized light fields to the alternative of using the appearance flow method~\cite{Zhou16}, a state-of-the-art view synthesis method that predicts a flow field to warp an input image to an image from a novel viewpoint. Other recent view synthesis methods are designed for predicting a held-out image from a stereo pair, so it is unclear how to adapt them to predict a 4D light field. On the other hand, it is straightforward to adapt the appearance flow method to synthesize a full 4D light field by modifying our depth estimation network to instead predict x and y flow fields to synthesize each sub-aperture image from the input view. We train this network on our training dataset. While appearance flow can be used to synthesize a light field, it does not produce any explicit geometry representation, so unlike our method, appearance flow cannot be used as a strategy for unsupervised geometry learning from light fields.

\begin{figure}
\begin{center}
\newcommand{\width}{1.0\linewidth}
\includegraphics[width=\width]{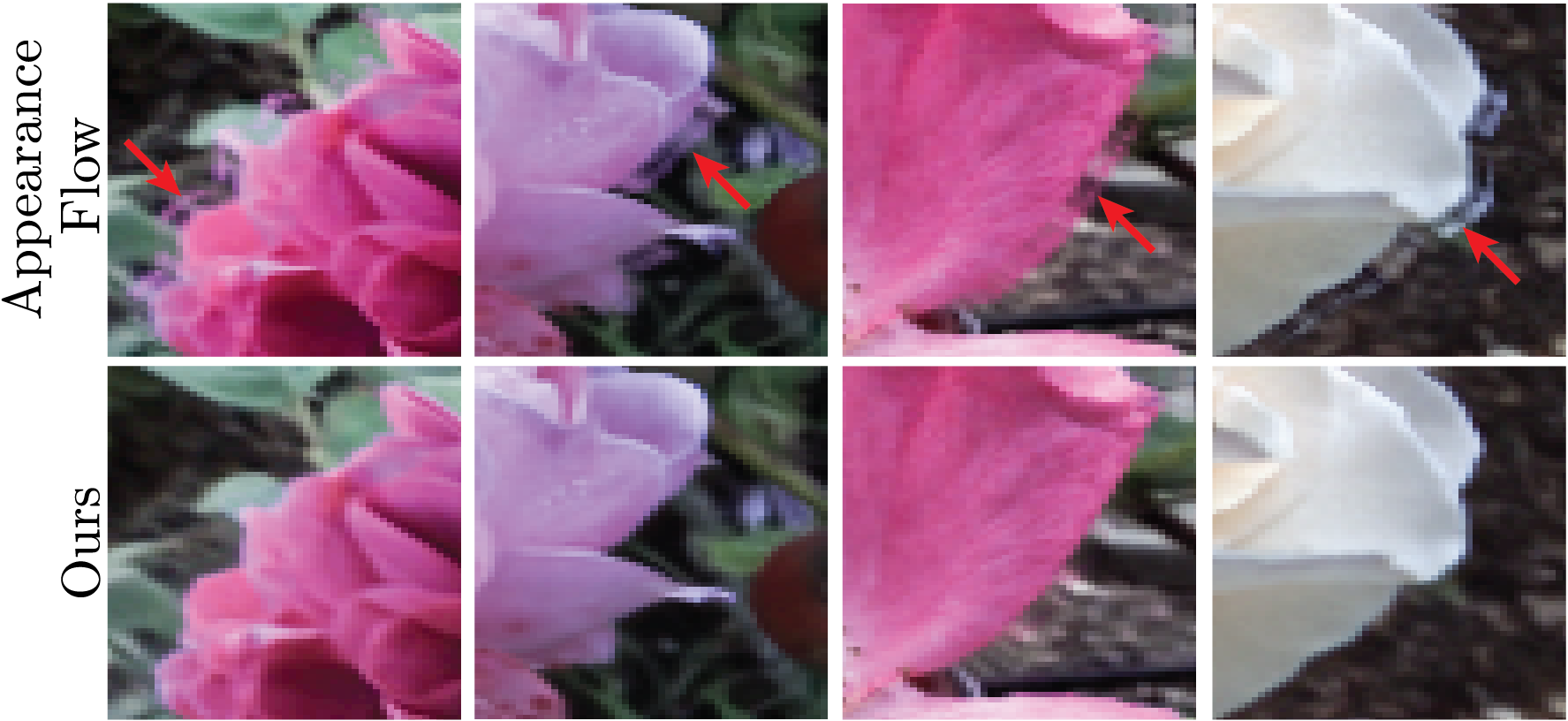} 
\caption{We compare our synthesized light fields to the appearance flow method~\cite{Zhou16}. Qualitatively, appearance flow has difficulties correctly predicting rays occluded from the input view, resulting in artifacts around the edges of the flowers. These types of edge artifacts are highly objectionable perceptually, and the improvement provided by our algorithm subjectively exceeds the quantitative improvement given in Fig.~\ref{fig:quantitative}.}
\label{fig:flow}
\end{center}
\vspace{-0.25in}
\end{figure}

Figure~\ref{fig:flow} illustrates that appearance flow has trouble synthesizing rays occluded from the input view, resulting in artifacts around occlusion boundaries. Our method is able to synthesize plausible occluded rays and generate convincing light fields. Intuitively, the correct strategy to flow observed rays into occluded regions will change dramatically for flowers with different colors and shapes, so it is difficult to learn. Our approach separates the problems of depth prediction and occluded ray prediction, so the depth prediction network can focus on estimating depth correctly without needing to correctly predict all occluded rays.

\begin{figure}
\begin{center}
\newcommand{\width}{1.0\linewidth}
\includegraphics[width=\width]{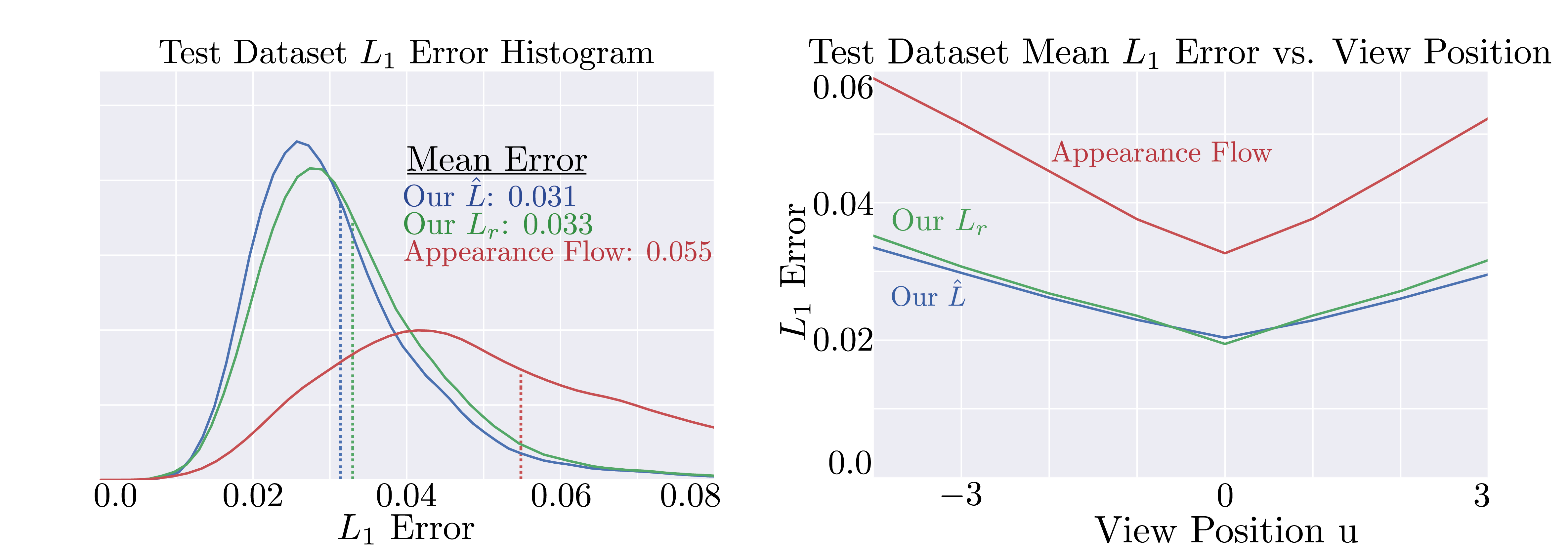} 
\caption{To quantitatively validate our results, we visualize histograms of the $L_1$ errors on the testing dataset for the outermost views of our predicted light fields $\hat{L}$, our Lambertian light fields $L_r$, and the light fields predicted by appearance flow. Our predicted light fields and Lambertian light fields both have lower errors than those of appearance flow. We also compute the mean $L_1$ errors as a function of view position $u$, and demonstrate that our algorithm consistently outperforms appearance flow.}
\label{fig:quantitative}
\end{center}
\vspace{-0.25in}
\end{figure}

To quantitatively evaluate our method, we display histograms for the mean $L_1$ error on our test dataset for our predicted light fields, our Lambertian light fields, and the appearance flow light fields in Fig.~\ref{fig:quantitative}. We calculate this error over the outermost generated views, since these are the most difficult to synthesize from a central input view. Our predicted light fields have the lowest mean error, and both our predicted and Lambertian approximate light fields have a lower mean error than the appearance flow light fields. We also plot the mean $L_1$ error as a function of the view position in $u$, and show that while all methods are best at synthesizing views close to the input view ($(u,v)=0$), both our predicted and Lambertian light fields consistently outperform the light fields generated by appearance flow. We also tested a CNN that directly regresses from an input image to an output light field, and found that our model outperforms this network with a mean $L_1$ error of 0.026 versus 0.031 across all views. Please refer to our supplementary material for more quantitative evaluation.

Encouragingly, our single view light field synthesis method performs only slightly worse than the light field interpolation method of~\cite{Kalantari16} that takes 4 corner views as input, with a mean L1 error of 0.0176 compared to 0.0145 for a subset of output views not input to either method.

\begin{figure}
\begin{center}
\newcommand{\width}{1.0\linewidth}
\includegraphics[width=\width]{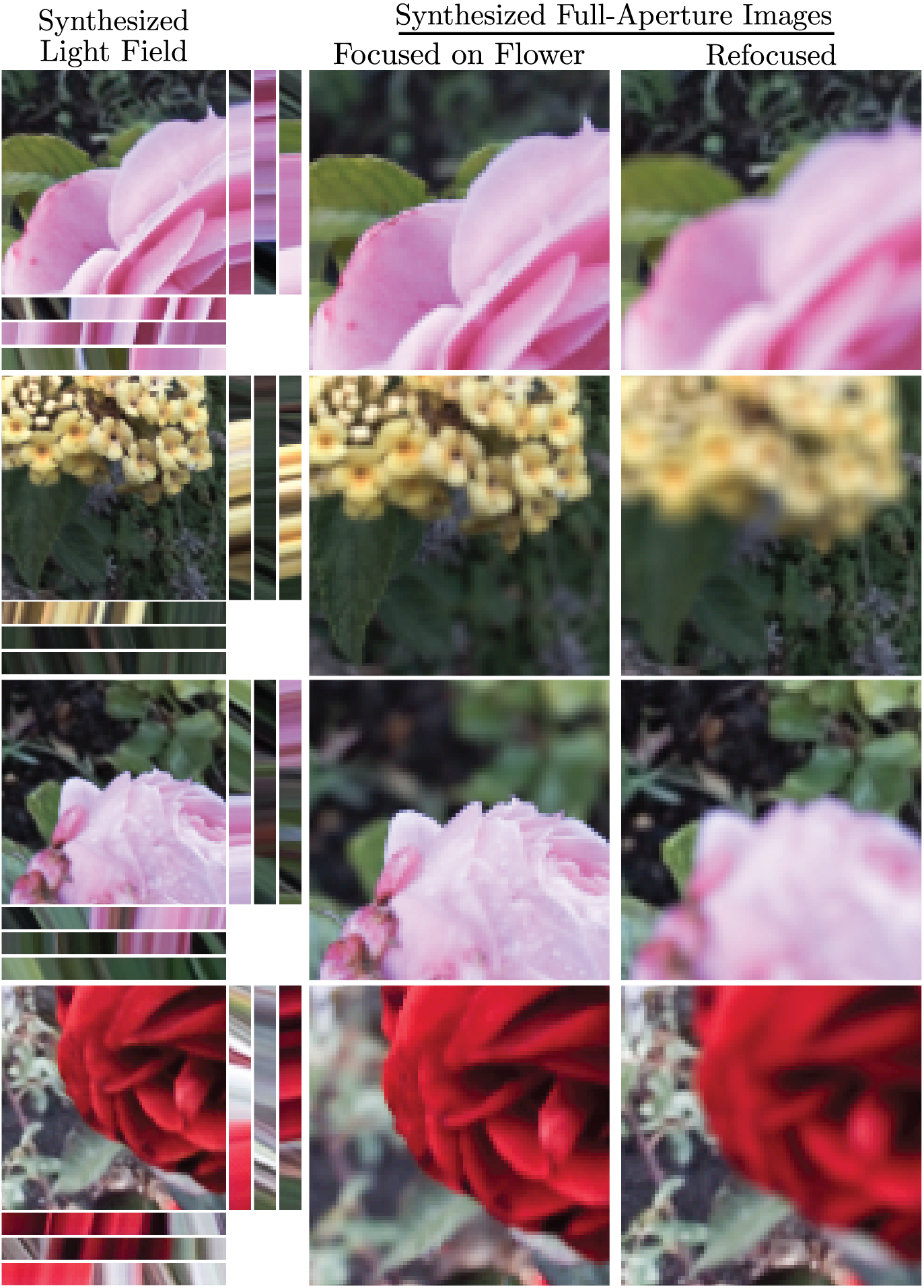} 
\caption{We visualize our synthesized light fields as a corner view crop, along with several epipolar slice crops. The epipolar slices demonstrate that our synthesized light fields contain complex occlusions and relative depths. We additionally demonstrate that our light field generated from a single 2D image can be used for synthetic defocus blur, increasing the aperture from $f/28$ to $f/3.5$. Moreover, we can use our light fields to convincingly refocus the full-aperture image from the flowers to the background.}
\label{fig:results}
\end{center}
\vspace{-0.25in}
\end{figure}

Figure~\ref{fig:results} displays example light fields synthesized by our method, and demonstrates that we can use our synthesized light fields for photographic effects. Our algorithm is able to predict convincing light fields with complex occlusions and depth ranges, as visualized in the epipolar slices. Furthermore, we can produce realistic photography effects, including extending the aperture from $f/28$ (aperture of the input view) to $f/3.5$ for synthetic defocus blur, and refocusing the full-aperture image from the flower to the background.

Finally, we note that inference is fast, and it takes under 1 second to synthesize a 187x270x8x8 light field and ray depths on a machine with a single Titan X GPU.

\begin{figure}
\begin{center}
\newcommand{\width}{1.0\linewidth}
\includegraphics[width=\width]{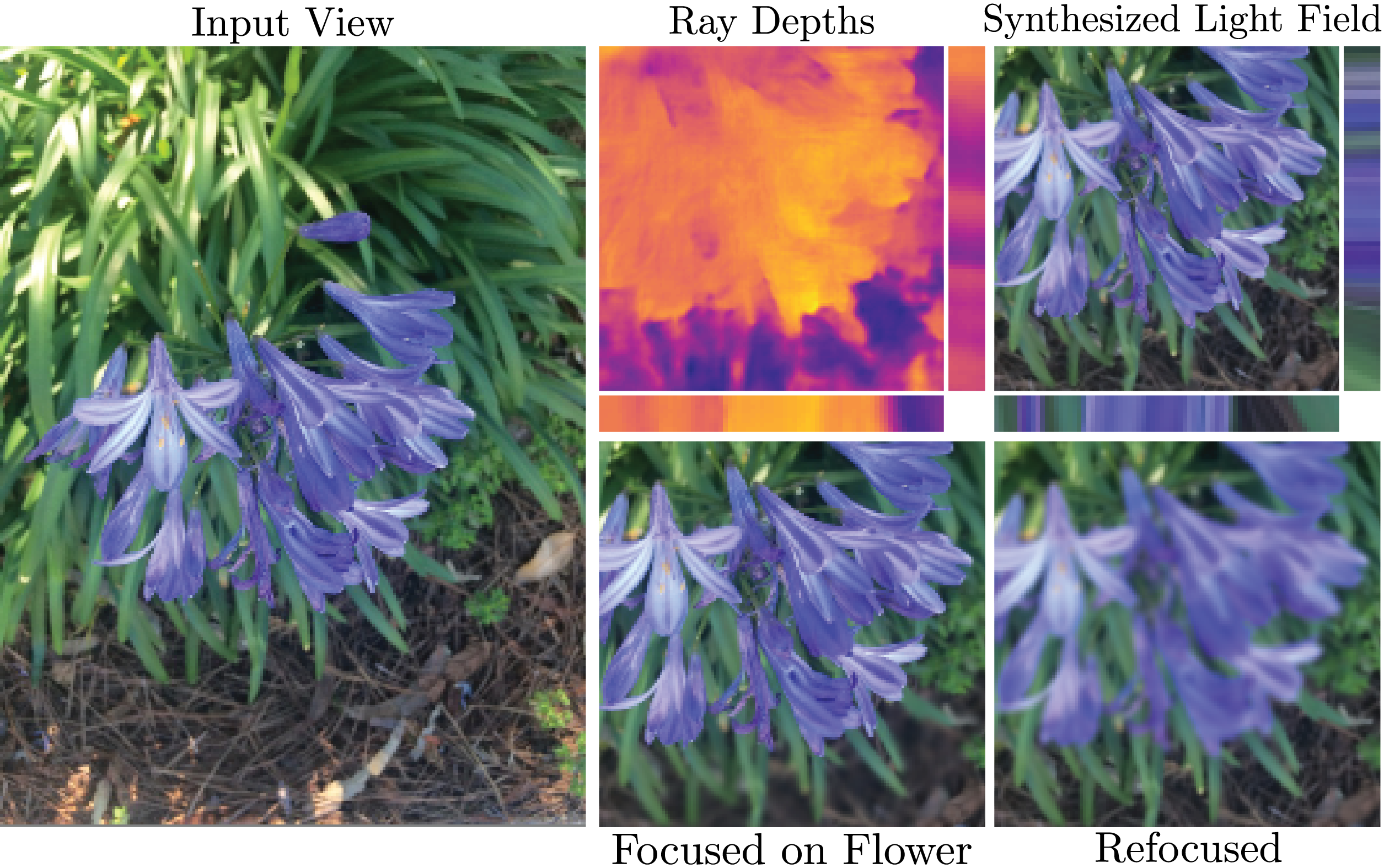} 
\caption{Our pipeline applied to cell phone photographs. We demonstrate that our network can generalize to synthesize light fields from pictures taken with an iPhone 5s. We synthesize realistic depth variations and occlusions, as shown in the epipolar slices. Furthermore, we can synthetically increase the iPhone aperture size and refocus the full-aperture image.}
\label{fig:cell_phone}
\end{center}
\vspace{-0.15in}
\end{figure}

\vspace{-0.15in}
\paragraph{Generalization} 
Figure~\ref{fig:cell_phone} demonstrates our method's ability to generalize to input images from a cell phone camera. We show that we can generate convincing ray depths, a high-quality synthesized light field, and interesting photography effects from an image taken with an iPhone 5s. 

\begin{figure}
\begin{center}
\newcommand{\width}{1.0\linewidth}
\includegraphics[width=\width]{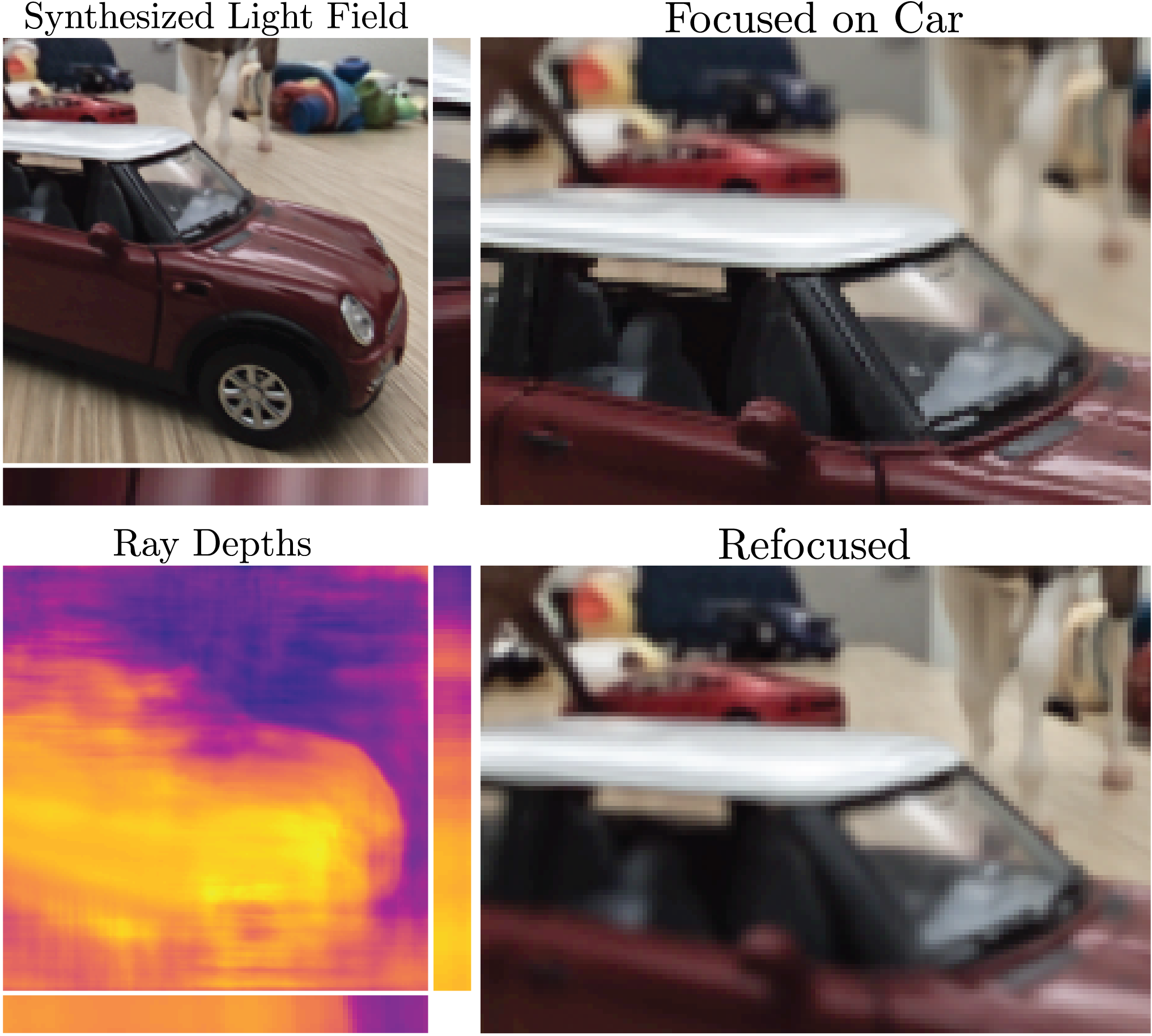} 
\caption{We demonstrate that our approach can generalize to scenes of toys, and we display an example test set result.}
\label{fig:toys}
\end{center}
\vspace{-0.25in}
\end{figure}

Finally, we investigate our framework's ability to generalize to other scene classes by collecting a second dataset, consisting of 4281 light fields of various types of toys including cars, figurines, stuffed animals, and puzzles. Figure~\ref{fig:toys} displays an example result from the test set of toys. Although our performance on toys is quantitatively similar to our performance on flowers (the mean $L_1$ error on the test dataset over all views is 0.027 for toys and 0.026 for flowers), we note that the toys results are perceptually not quite as impressive. The class of toys is much more diverse than that of flowers, and this suggests that a larger and more diverse dataset would be useful for this scene category. 

\vspace{-0.05in}
\section{Conclusion}
\vspace{-0.05in}

We have shown that consumer light field cameras enable the practical capture of datasets large enough for training machine learning algorithms to synthesize local light fields of specific scenes from single photographs. It is viable to extend this approach to other niches, as we demonstrate with toys, but it is an open problem to generalize this to the full diversity of everyday scenes. We believe that our work opens up two exciting avenues for future exploration. First, light field synthesis is an exciting strategy for unsupervised geometry estimation from a single image, and we hope that our dataset and algorithm enable future progress in this area. In particular, the notion of enforcing consistent geometry for rays that intersect the same scene point can be used for geometry representations other than ray depths, including voxels, point clouds, and meshes. Second, synthesizing dense light fields is important for capturing VR/AR content, and we believe that this work enables future progress towards generating immersive VR/AR content from sparsely-sampled images. 

\vspace{-0.1in}
\paragraph{Acknowledgments} This work was supported in part by ONR grants N00014152013 and N000141712687, NSF grant 1617234, NSF fellowship DGE 1106400, a Google Research Award, the UC San Diego Center for Visual Computing, and a generous GPU donation from NVIDIA.

\clearpage
{\small
\bibliographystyle{ieee}
\bibliography{PRATUL}
}

\end{document}